\pdfoutput=1

\documentclass[11pt]{article}

\usepackage[review]{acl}

\usepackage{times}
\usepackage{latexsym}
\usepackage{ulem}
\usepackage{amsmath}
\usepackage{stfloats}
\usepackage{graphicx}
\usepackage{booktabs}
\usepackage{multirow}
\usepackage{multicol}

\usepackage[T1]{fontenc}

\usepackage[utf8]{inputenc}

\usepackage{microtype}

\title{Learning to Rank Utterances for Query-Focused Meeting Summarization}

\author{Xingxian Liu, Yajing Xu\footnotemark[1] \\
  Pattern Recognition \& Intelligent System Laboratory \\
  Beijing University of Posts and Telecommunications, Beijing, China \\
  \texttt{\{liuxingxian,xyj\}@bupt.edu.cn} \\}
\begin{document}
\maketitle

\renewcommand{\thefootnote}{\fnsymbol{footnote}} 
\footnotetext[1]{Yajing Xu is the corresponding author.}

\begin{abstract}
Query-focused meeting summarization(QFMS) aims to generate a specific summary for the given query according to the meeting transcripts. 
Due to the conflict between long meetings and limited input size,
previous works mainly adopt extract-then-summarize methods, which use extractors to simulate binary labels or ROUGE scores to extract utterances related to the query and then generate a summary.
However, the previous approach fails to fully use the comparison between utterances. To the extractor, comparison orders are more important than specific scores.
In this paper, we propose a \textbf{Ranker-Generator} framework. 
It learns to rank the utterances by comparing them in pairs and learning from the global orders, then uses top utterances as the generator's input.
We show that learning to rank utterances helps to select utterances related to the query effectively, and the summarizer can benefit from it.
Experimental results on QMSum show that the proposed model outperforms all existing multi-stage models with fewer parameters.

\end{abstract}

\section{Introduction}
Query-focused meeting summarization(QFMS) aims to summarize the crucial information for the given query into a concise passage according to the meeting transcripts. 
By responding to the query, QFMS can meet the user's need to focus on a specific aspect or topic of the meeting \cite{litvak-vanetik-2017-query, baumel2018query}.
Unlike the generic summary, QFMS requires the summary depending on both the given query and meeting transcripts. 

Previous works consist of end-to-end and two-stage frameworks. 
The end-to-end models take the whole long meeting as the input. 
Although some works such as HMNet \cite{zhu-etal-2020-hierarchical} and HATBART \cite{Rohde2021HierarchicalLF} use hierarchical attention mechanism to alleviate the rapid growth in computational complexity, it's still faced with difficulties in training efficiency.
The two-stage models extract utterances related to the query and then pass the concatenation of them to the generator.
For QFMS, the key information related to the query scatters in certain parts of the meeting.
Therefore, the two-stage framework is considered as a practical approach to balance experimental performance and computational efficiency in the long-input problems.

\begin{figure}[t]
	\centering
    \includegraphics[scale=0.68]{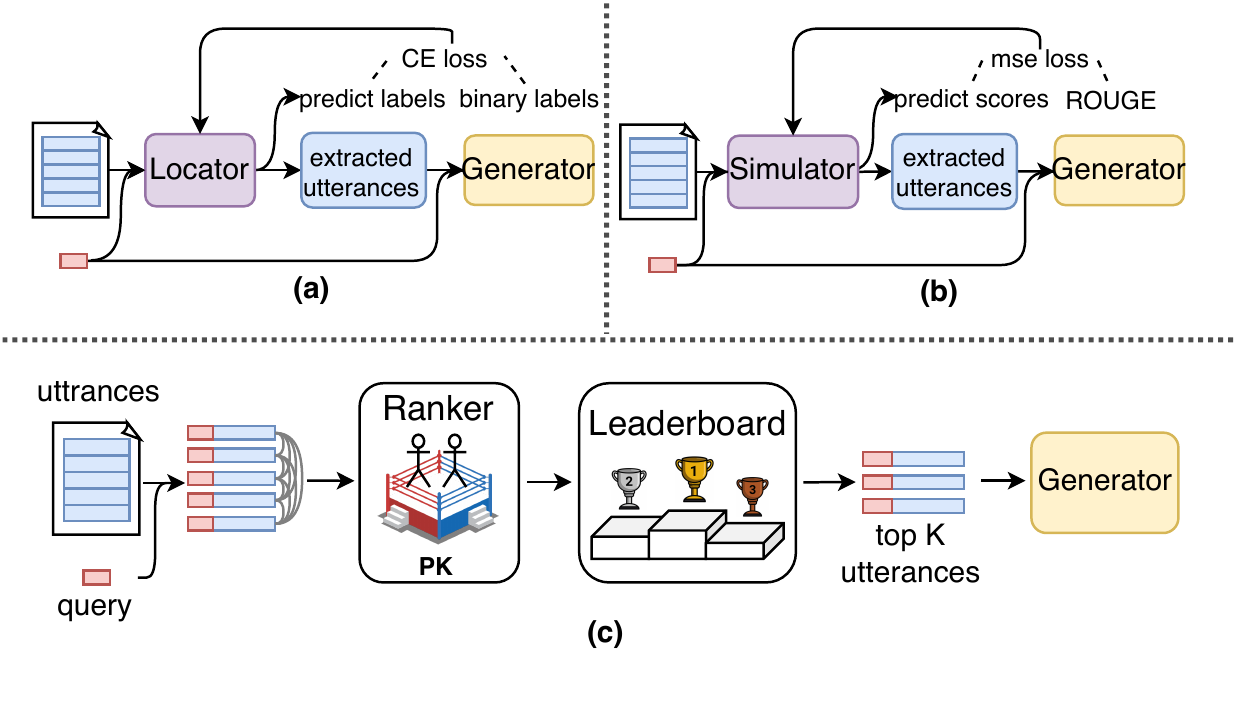}%
	\caption{
        (a) \textbf{Locator-Generator} framework, it predicts a binary label and uses Cross-Entropy loss to update parameters.
        (b) \textbf{Simulator-Generator} framework, it simulates the ROUGE score and uses Mean Squared Error loss to update parameters.
        (c) \textbf{Ranker-Generator} framework proposed in this paper, it learns to rank utterances from the relative order between utterances. The top K utterances can be passed to the generator.
	}
	\label{fig1}
\end{figure}

\begin{figure*}[t]
	\centering
    \includegraphics[scale=0.78]{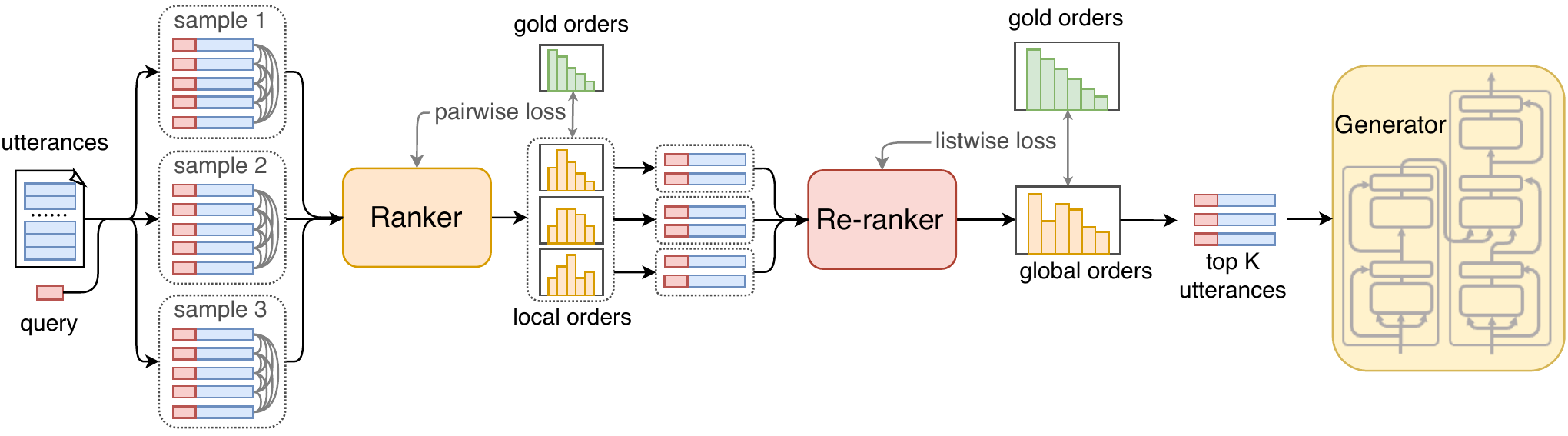}%
	\caption{The overall model structure}
	\label{fig2}
\end{figure*}

The two-stage framework mainly includes the Locator-Generator and the Simulator-Generator approaches. 
As shown in Figure \ref{fig1}, in the first stage, the Locator-Generator \cite{zhong-etal-2021-qmsum} framework considers it as a binary classification task. It predicts a binary label of whether the utterance is relevant to the query and uses cross-entropy loss to update parameters. But the hard binary labels can not reflect the relative quality. Especially when the training data is limited by scarcity, the binary classification will have a large margin between positive and negative samples. So the Simulator-Generator \cite{vig-etal-2022-exploring} framework considers it as a ROUGE score regression task. It simulates the ROUGE score and uses MSE loss to update parameters. 
However, there is a gap between the extractor's ultimate objective and the objective of minimizing the absolute error between predicted scores and ROUGE scores. In fact, rather than specific scores, we care more about the relative orders of utterances.

To make full use of the comparison information between samples, we propose a Ranker-Generator framework in this paper. 
To balance experimental effectiveness and computational efficiency, the framework contains three steps.
First, the utterances would be divided into samples. We conduct pairwise ranking to get an order for each sample. 
Second, the top utterances in different samples would be fed into the re-ranker, which would conduct listwise ranking to get a global order.
Finally, the top K utterances would be concatenated and passed to the generator.

To summarize, our contributions are as follows:
(1) This paper demonstrates that, by enhancing the accuracy of extracting query-relevant utterances, the generator can make the summary more related to the query.
(2) We propose a Ranker-Generator framework to extract query-relevant utterances by learning to rank discourse to improve the quality of the generated summaries.
(3) Experimental results show that the proposed model outperforms existing multi-stage models with fewer model parameters.


\section{Method}
The architecture of our method is illustrated in Figure \ref{fig2}. 
Our model consists of a two-stage ranking step and a generating step. 
The utterances would be ranked by the Sample Pairwise Ranking module and the Global Listwise Re-ranking module, and top of them can be passed to the generator to produce the final summary.

\subsection{Two-Stage Ranking}
The utterance ranking orders for a brief meeting can be efficiently obtained using the single-stage ranking paradigm. 
However, the computing complexity of full-pairwise ranking grows at a square rate as the number of utterances grows.
Therefore, we adopt a two-stage ranking framework.
In the first stage, we propose sample pairwise ranking to reduce computational complexity.
But sample pairwise ranking can only evaluate the relative quality within samples. 
It performs poorly when applied to utterances from various samples, e.g., the top utterances in sample 1 may be ranked lower in sample 2.
To overcome the above problem, we apply global listwise re-ranking and concentrate on the top-k utterances in the second stage.
Utterances that are unlikely to appear in the generator are filtered out by the pairwise ranking model, then global listwise ranking is conducted to get better top-k orders.

\subsection{Sample Pairwise Ranking}
In this paper, the ROUGE \cite{lin-2004-rouge} scores between utterances $U$ and the gold summary $S^*$ are considered as the measure of query-relevance. 
The utterances from one meeting are divided into various samples. In one sample, the utterances would be ordered by the ROUGE scores. 
The ranker should be encouraged to assign higher relevance scores to these top utterances in the order. 
By learning to rank in pairwise, the model can distinguish the utterances that are more relevant to the query from the comparison.
Following the previous work \cite{zhong-etal-2020-extractive}, the loss is as follows:
\begin{equation}
    L=\sum_{i}{\sum_{j>i}{max(0, f(U_j)-f(U_i)+\lambda_{ij})}}
\end{equation}
\vspace{-5mm}
\begin{equation}
    \lambda_{ij} = (j-i) * \lambda
\label{equ2}
\end{equation}
where $U_i$ and $U_j$ are the $i$-th and $j$-th utterances in gold ranking orders, ROUGE($U_i, S^{\ast}$)$>$ROUGE($U_j, S^{\ast}$), $\forall i, j, i < j$,  $\lambda$ is the base margin. $f(U_i)$ is the predicted query-relevance score given by a cross-encoder model.


\subsection{Global Listwise Re-ranking}
As shown in Figure \ref{fig2}, the top utterances in different samples are gathered in the re-ranking module. 
The gold orders would be determined by ranking the utterances according to the ROUGE scores. 
To obtain a more precise top-ranking order, we would perform a refined global sort on these top utterances from various samples using listwise re-ranking.
Inspired by ListNet \cite{Cao2007LearningTR}, we optimize the permutation probability distribution between predicted scores $s$ and the gold scores $s^{\ast}$.
The permutation probability is defined as
\begin{equation}
    P_s(\pi)=\prod_{j=1}^n\frac{\phi(s_{\pi(j)})}{\sum^n_{t=j}{\phi(s_{\pi(t)})}}
\end{equation}
$\pi$ is a permutation on the n objects, and $\phi(.)$ is an increasing and strictly positive function. 

But different with ListNet, we optimize the top-k permutation probability rather than top-1 probability.
The top-k permutation probability is as follows:
\begin{equation}
    P_s^k(\pi)=\prod_{j=1}^k\frac{\phi(s_{\pi(j)})}{\sum^n_{t=j}{\phi(s_{\pi(t)})}}
\end{equation}
For example, the top-3 permutation probability of $\pi=\langle 1,2,3,4,5 \rangle$ is as follows:
\begin{equation}
\resizebox{0.96\columnwidth}{!}{
    $P_s^{3}(\pi)=\frac{\phi(s_1)}{\sum_{i=1}^5{\phi(s_i)}} \cdot \frac{\phi(s_2)}{\sum_{i=2}^5{\phi(s_i)}} \cdot \frac{\phi(s_3)}{\sum_{i=3}^5{\phi(s_i)}}$
}
\end{equation}
The predicted top1-to-topk distribution is $P_{s}=(P_s^{1}, P_s^{2},\cdots,P_s^{k})$, the gold top1-to-topk distribution is $P_{s^\ast}=(P_{s^*}^{1}, P_{s^*}^{2},\cdots,P_{s^*}^{k})$
We use KL-divergence to reduce the gap between the above two distributions.
\vspace{-1mm}
\begin{equation}
    L=KL(P_{s^\ast}||P_{s})
\end{equation}
\vspace{-5mm}
\begin{equation}
    KL(P_{s^\ast}||P_{s})=\sum_{i=1}^{k}{P^i_{s^{\ast}}\cdot \log{\frac{P^i_{s^{\ast}}}{P^i_{s}}}}
\end{equation}






\subsection{Generator}
As shown in Figure \ref{fig2}, after the two-stage ranking, top-k of the utterances would be concatenated and fed into the generator. In the generation stage, the objective is to minimize the cross-entropy loss:
\begin{equation}
    L=-\sum_{i}{p_{gt}(S_i|S^{\ast}_{<i},U)\log{p(S_i|S^{\ast}_{<i},U)}}
\end{equation}

\begin{equation}
    p_{gt}(S_i|S^{\ast}_{<i},U)=
    \begin{cases}
        1& S_i=S_i^{\ast}\\
        0& S_i\neq S_i^{\ast}
    \end{cases}
\end{equation}
$U$ is the generator's input, $S^{\ast}$ is the gold summary.

\section{Experiments}


\begin{table*}
\centering
\resizebox{1.70\columnwidth}{!}{
\begin{tabular}{lcccc}
\specialrule{0.13em}{2pt}{2pt}
\textbf{Models} & \textbf{ROUGE-1} & \textbf{ROUGE-2} & \textbf{ROUGE-L} & \textbf{Extractor Size(M)}\\
\specialrule{0.075em}{1pt}{1pt}
TextRank \cite{mihalcea-tarau-2004-textrank}    &   16.27   &   2.69    &   15.41   &   -   \\
PGNet \cite{see-etal-2017-get}       &   28.74   &   5.98    &   25.13   &   -   \\
BART \cite{lewis-etal-2020-bart}            &   29.20   &   6.37    &   25.49   &  -    \\
LEAD + BART &   32.06   &   9.67    &   27.93   & - \\
HMNet \cite{zhu-etal-2020-hierarchical}  &   32.29   &   8.67    &   28.17   &  -   \\
Longformer \cite{Beltagy2020Longformer} & 34.18 & 10.32 & 29.95 & - \\
DialogLM \cite{Zhong2021DialogLMPM} &   33.69   &   9.32    &   30.01   &   -   \\
SUMM$^N$ \cite{zhang-etal-2022-summn}                   &    34.03  &   9.28    &   29.48   &   -   \\
DYLE \cite{mao-etal-2022-dyle}       &  34.42   &   9.71    &   30.10   &   501   \\
Pointer Network + PGNet \cite{zhong-etal-2021-qmsum}  &   31.37   &   8.47    &   27.08   &   440 \\
Pointer Network + BART \cite{zhong-etal-2021-qmsum}   &   31.74   &   8.53    &   28.21   &   440 \\
RELREG-TT \cite{vig-etal-2022-exploring}    &   33.02   &   10.17   &  28.90   &   329 \\
RELREG \cite{vig-etal-2022-exploring}    &   34.91   &   11.91   &  30.73   &   1372 \\
\specialrule{0.075em}{1pt}{1pt}
Oracle  &   43.80   &   19.63   &   39.10   \\
Locator-Generator   &       31.47(-3.77)    &   8.53(-3.70) &   28.21(-3.07)    &   134 \\
Simulator-Generator &   32.92(-2.59)    &   9.46(-2.77) &   28.93(-2.35)    &   134 \\
\textbf{Ranker-Generator}   &   \textbf{35.51}   &   \textbf{12.23}   &   \textbf{31.28}   &   134   \\
RankSUM(w/o re-ranking)  &   33.02(-2.49)   &   9.73(-2.50)    &   29.15(-2.13)   & 134   \\
\specialrule{0.13em}{2pt}{2pt}
\end{tabular}
}
\caption{
\label{tab1}
    ROUGE-F1 scores for different models on QMSum dataset.
}
\end{table*}

\subsection{Setup}
\subsubsection{Implementation Details}
Models are implemented using the PyTorch framework. 
The pre-trained BART\footnote{The checkpoint is “facebook/bart-large”, containing around 400M parameters.} from the Transformers \cite{wolf-etal-2020-transformers} library is used as the base abstractive model.
The pre-trained MiniLM\footnote{The checkpoint is “cross-encoder/ms-marco-MiniLM-L-12-v2”, containing around 134M parameters.} from the sentence-transformers \cite{reimers-gurevych-2019-sentence} library is used as the pairwise ranking model and the listwise re-ranking model.

All experiments are conducted on NVIDIA RTX 3090 GPU(24G memory). 
The generator model is trained for 10 epochs. For one model training, the average running time is around 2 hours.
Weight hyperparameter $\lambda$ is 0.01 in Equation \ref{equ2}. The generator's max length of the input is 1024, max length of the output is 256. 
Learning rate is 5e-6.

Models were evaluated using the ROUGE metrics \cite{lin-2004-rouge} in the SummEval toolkit \cite{fabbri-etal-2021-summeval} and each pair of results was subjected to t-test to confirm the effectiveness of our method.

\subsubsection{Datasets Details}
\textbf{QMSum} \cite{zhong-etal-2021-qmsum} is a query-focused meeting summarization dataset consisting of 1,808 query-summary pairs over 232 meetings from product design, academic, and political committee meetings. 
Additionally, QMSum contains manual annotations such as topic segmentation and relevant spans related to the reference summary.

\subsubsection{Baselines Details}
We compare the proposed method with several baselines. 
\textbf{TextRank} \cite{mihalcea-tarau-2004-textrank} is an extractive summarization method with a graph-based ranking model. 
\textbf{PGNet} \cite{see-etal-2017-get} uses pointer mechanism to copy tokens from source texts.
\textbf{BART} \cite{lewis-etal-2020-bart} is a pre-trained encoder-decoder Transformer model with a denoising objective, which achieves advanced performance on several summarization datasets(i.e. CNN/DailyMail \cite{hermann2015teaching} and Xsum \cite{narayan-etal-2018-dont}).
\textbf{LEAD+BART} uses the beginning utterances as the BART's input.
\textbf{HMNet} \cite{zhu-etal-2020-hierarchical} uses a hierarchical attention mechanism and cross-domain pre-training for meeting summarization.
\textbf{Longformer} \cite{Beltagy2020Longformer} replaces the quadratic self-attention mechanism with a combination of local attention and sparse global attention.
\textbf{DialogLM} \cite{Zhong2021DialogLMPM} is a pre-train model using intra-window denoising self-reconstruction pre-training task and intra-block inter-block mixing attention. 
\textbf{SUMM$^N$} \cite{zhang-etal-2022-summn} is a multi-stage summarization framework for the long-input summarization task.
\textbf{DYLE} \cite{mao-etal-2022-dyle} treats the extracted text snippets as the latent variable and jointly trains the extractor and the generator.
\textbf{Point Network+PGNet} and \textbf{Point Network+BART} \cite{zhong-etal-2021-qmsum} adopt a two-stage approach of locate-then-summarize for long meeting summarization. 
\textbf{RELREG-TT} \cite{vig-etal-2022-exploring} and \textbf{RELREG} \cite{vig-etal-2022-exploring} considers extracting as a ROUGE regression model using bi-encoder and cross-encoder.

\begin{table}
\centering
\resizebox{0.90\columnwidth}{!}{
\begin{tabular}{l|ccc|ccc}
\toprule
\multirow{2}{*}{\textbf{Models}} & \multicolumn{3}{c|}{\textbf{Top 5}} & \multicolumn{3}{c}{\textbf{Top 10}} \\
& \textbf{R-1} & \textbf{R-2} & \textbf{R-L} & \textbf{R-1} & \textbf{R-2} & \textbf{R-L} \\
\hline
Gold & 26.32 & 7.58 & 24.43 & 20.55 & 5.15 & 19.29 \\
LEAD & 11.15 & 0.99 & 10.17 & 12.11 & 1.11 & 11.10 \\
RELREG & 18.02 & 2.46 & 15.30 & 15.02 & 2.35 & 13.23 \\
Locator & 16.89 & 2.24 & 13.97 & 14.10 & 1.97 & 12.75 \\ 
Simulator & 17.06 & 2.36 & 14.88  & 14.44 & 2.14 & 13.06 \\
Ours & \textbf{20.07} & \textbf{3.69} & \textbf{17.78}  & \textbf{17.08} & \textbf{3.01} & \textbf{15.48} \\
\bottomrule
\end{tabular}
}
\caption{
\label{tab2}
ROUGE-F1 scores between the gold summary and top-5/top-10 utterances for different models on QMSum.
}
\end{table}

\subsection{Results \& Analysis}
The ROUGE score \cite{lin-2004-rouge} is adopted as the evaluation metric.
The performances of our method and baselines are summarized in Table \ref{tab1}.
Experimental results show that our method significantly outperforms the baselines (p < 0.05) on QMSum dataset with fewer parameters.



To have a fair comparison among the three frameworks, we design an experiment to evaluate the performance of these frameworks using the same backbone as the extractor and the same generator.
The experimental results show that the proposed model significantly outperforms Locator-Generator and Simulator-Generator, which demonstrates that the ranker can obtain meeting utterances that are more suitable for the generator by learning to rank utterances.

To verify the effectiveness of the two-stage ranking paradigm, we conduct an ablation experiment.
Our model significantly outperforms the model without re-ranking module (p < 0.05).
Experimental results show that the model without re-ranking module reduces 2.49 ROUGE-1, 2.50 ROUGE-2, 2.13 ROUGE-L scores, which demonstrates the importance of the re-ranking module. By listwise ranking, we can get a more precise top-ranking order.

We have an interesting observation. 
Unlike the ROUGE score regression model, the ranker is less sensitive to the model size.
We believe this is because learning the relative order by comparison is easier than fitting ROUGE scores separately.
It reduces the ranker's reliance on the model size by making full use of the comparison between samples.
As a training task for extractors, learning to rank is a more suitable objective.
Since to the extractor, it is the relative order that matters rather than the absolute error in fitting the ROUGE score.

\subsection{Extractor Performance}
We conduct experiments to evaluate the performance of the extractor, which help to explore the impact of the extractor on the quality of the generated summaries.
The lexical overlap metric between the extracted utterances and the gold summary is used to measure the relevance of the meeting utterances to the summary/query.
The experimental results show that the ranker significantly outperforms the baselines in extracting relevant utterances.
It demonstrates that by learning to rank utterances, the ranker is able to extract the utterances that are more relevant to the summary/query.

\subsection{Human Evaluation}
We further conduct a manual evaluation to assess the models. We randomly select 50 samples from QMSum and ask 5 professional linguistic evaluators to score the ground truth and summaries generated by 5 models according to 3 metrics: fluency, query relevance and factual consistency. 
Each metric is rated from 1 (worst) to 5 (best) and the scores for each summary are averaged.

As shown in Table 3, the proposed model significantly outperforms all the baselines on query relevance, which benefits from the extractor’s improvement on selecting the relevant utterances. 
Besides, the factual consistency score is also improved. 
We think that by comparing the relevance between utterances and the summary/query, the top utterances are more relevant to each other, which may help to improve factual consistency.
In the aspect of fluency, the proposed model has only slight improvement compared to the baselines.

\begin{table}
\centering
\resizebox{0.75\columnwidth}{!}{
\begin{tabular}{lccc}
\toprule
\textbf{Models} & \textbf{Flu.} & \textbf{QR.} & \textbf{FC.} \\
\midrule
Gold & 4.88 & 4.90 & 4.92 \\
BART & 4.48 & 3.78 & 3.64 \\
RELREG & 4.51 & 4.12 & 4.07 \\
Locator-Generator & 4.45 & 3.90 & 3.83 \\ 
Simulator-Generator & 4.48 & 4.01 & 4.02  \\
Ours & \textbf{4.52} & \textbf{4.40} & \textbf{4.21} \\
\bottomrule
\end{tabular}
}
\caption{
\label{tab3}
Human evaluation on Fluency (Flu.), Query Relevance (QR.) and Factual Consistency (FC.) for QMSum.
}
\end{table}




\section{Conclusion}
This paper proposes a new multi-stage framework for QFMS. It learns to rank the meeting utterances by pairwise and listwise comparison between them. By selecting the utterances with high query-relevance scores as the generator’s input, the generator can produce high-quality summaries that are more relevant to the query. The experiments demonstrate the effectiveness of the Ranker-Generator framework.

\section{Acknowledgements}
This work was supported by MoE-CMCC ”Artiﬁcal Intelligence” Project No. MCM20190701 and the National Natural Science Foundation of China (NSFC No.62076031).

We thank the anonymous reviewers for valuable feedback and helpful suggestions.

\section*{Limitations}
This paper mainly focuses on the Query-focused Meeting Summarization(QFMS) task. 
Besides, We have explored the performance of the Ranker-Generator framework on the long-input summarization task.
But the results do not show a significant improvement. 
Although QMSum dataset is also faced with the long-input challenge, the QFMS task only summarizes specific parts of the original text, so it can take these parts as the input. While the goal of the long-input summarization task is to generate an overall summary, which needs to have a global view on the original text. So we think the extract-then-generate framework is unsuitable for the long-input summarization task. The previous work SUMM$^N$ \cite{zhang-etal-2022-summn} is more suitable for the long-input summarization task.

In addition, the multi-stage approach has a performance disadvantage over the end-to-end approach.
However, the computational complexity of the multi-stage approach is much lower than that of the end-to-end approach. 
The multi-stage approach can balance experimental performance and computational complexity.
So it is worthy of exploration as well as the end-to-end approach.


\section*{Ethics Statement}
In this paper, all experiments are conducted on \textbf{QMSum} \cite{zhong-etal-2021-qmsum}, which is open-source and obeys MIT license. The meeting transcripts data doesn't contain any privacy information(such as password, phone number and trade secrets) or offensive content.

\bibliography{anthology,custom}

\newpage
\appendix

\section{Case Study}
\label{sec:appendixa}

\begin{figure}[hb]
	\centering
    \includegraphics[scale=0.72]{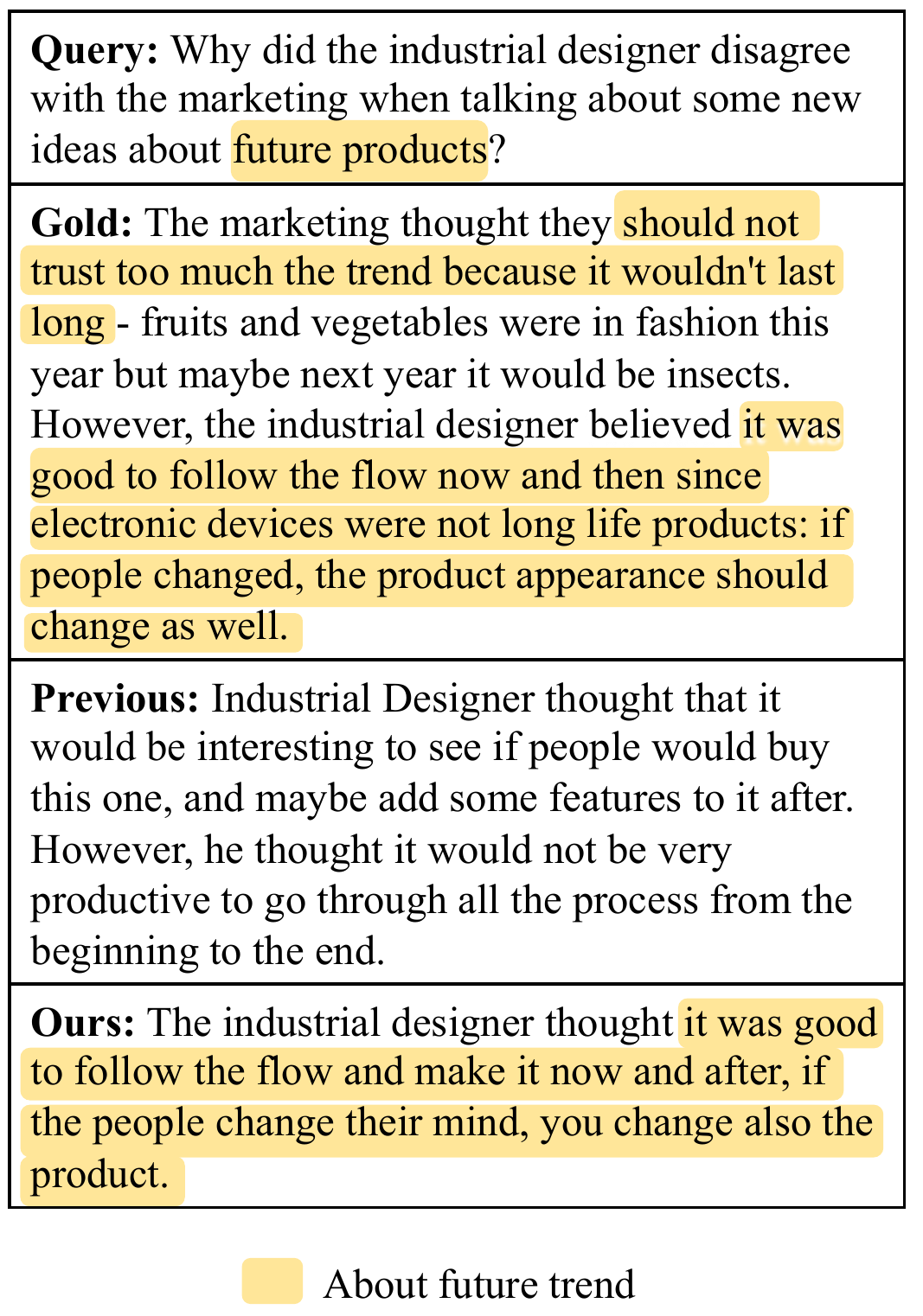}%
	\label{fig3}
\end{figure}

\clearpage

\begin{figure*}[tp]
	\centering
    \includegraphics[scale=0.70]{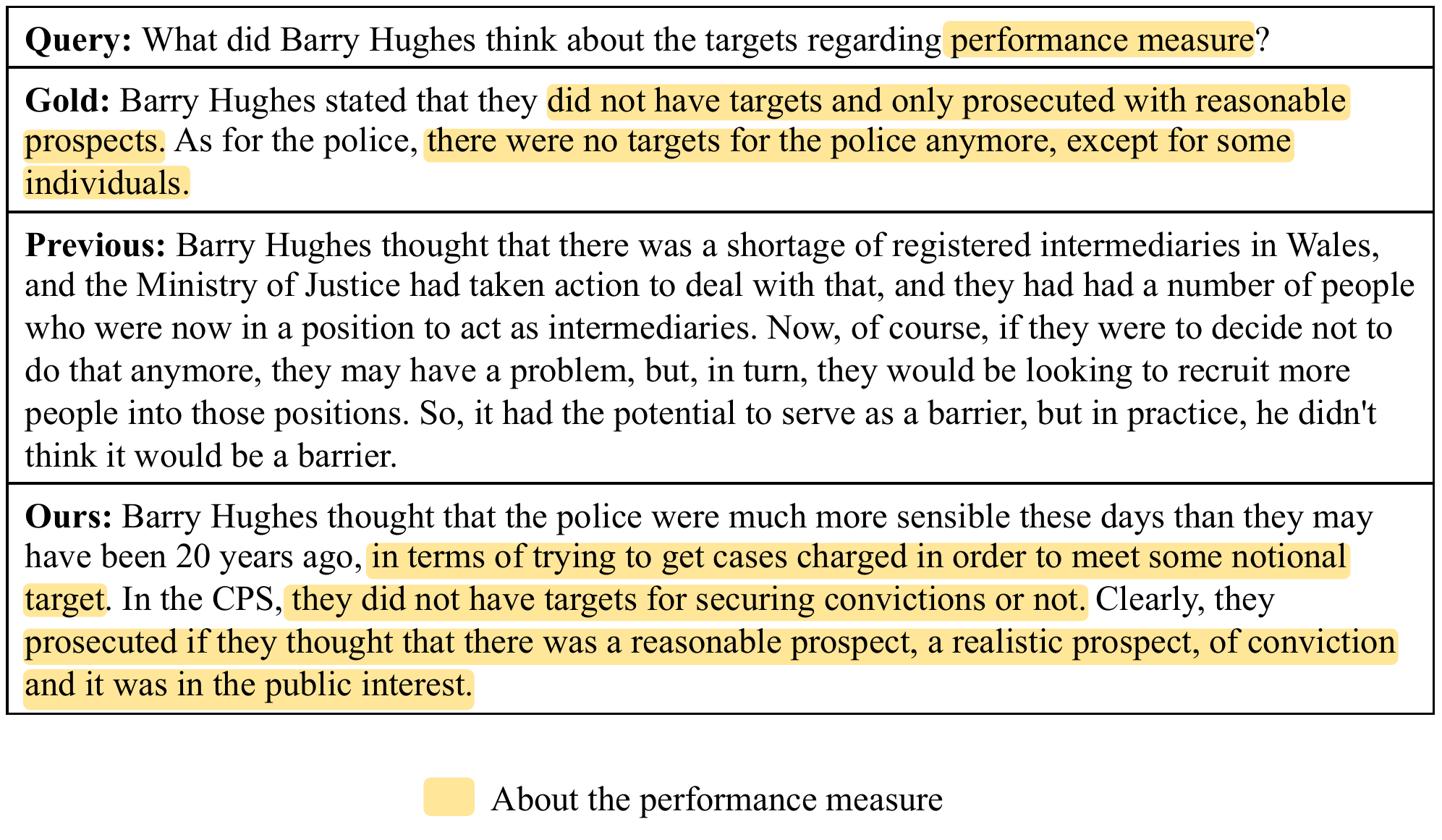}%
	\label{fig4}
\end{figure*}

\end{document}